# More Effective Crossover Operators for the All-Pairs Shortest Path Problem[*]

Benjamin Doerr[†], Daniel Johannsen[‡][§], Timo Kötzing [†][¶]
Frank Neumann[‖][**], Madeleine Theile[††]

June 4, 2018

## Abstract

The all-pairs shortest path problem is the first non-artificial problem for which it was shown that adding crossover can significantly speed up a mutation-only evolutionary algorithm. Recently, the analysis of this algorithm was refined and it was shown to have an expected optimization time (w. r. t. the number of fitness evaluations) of $\Theta(n^{3.25}(\log n)^{0.25})$.

In contrast to this simple algorithm, evolutionary algorithms used in practice usually employ refined recombination strategies in order to avoid the creation of infeasible offspring. We study extensions of the basic algorithm by two such concepts which are central in recombination, namely *repair mechanisms* and *parent selection*. We show that repairing infeasible offspring leads to an improved expected optimization time of $O(n^{3.2}(\log n)^{0.2})$. As a second part of our study we prove that choosing parents that guarantee feasible offspring results in an even better optimization time of $O(n^3 \log n)$.

Both results show that already simple adjustments of the recombination operator can asymptotically improve the runtime of evolutionary algorithms.

[*]This work greatly benefited from various discussions at the Dagstuhl seminar 08051 on Theory of Evolutionary Algorithms.
[†]Department 1: Algorithms and Complexity, Max-Planck-Institut für Informatik, 66123 Saarbrücken, Germany
[‡]School of Mathematical Sciences, Tel Aviv University, Tel Aviv 69978, Israel
[§]Part of this work was done while Daniel Johannsen was with the Max-Planck-Institut für Informatik, Saarbrücken. He is now supported by a fellowship within the Postdoc-Programme of the German Academic Exchange Service (DAAD).
[¶]Timo Kötzing was supported by the Deutsche Forschungsgemeinschaft (DFG) under grant no. NE 1182/5-1.
[‖]School of Computer Science, The University of Adelaide, Adelaide, SA 5005, Australia
[**]This work was partially done while Frank Neumann was affiliated with the Max-Planck-Institut für Informatik, Saarbrücken.
[††]Institut für Mathematik, Technische Universität Berlin 10623 Berlin, Germany

# 1  Introduction

Evolutionary algorithms [14] have been shown to be robust problem solvers for a wide range of combinatorial optimization problems that cannot be handled by traditional algorithmic approaches [34]. They are a premier choice for complex optimization problems that are highly non-linear, dynamic, and/or stochastic. Problems that can be observed in the real-world have most of these characteristics. Thus, it is desirable to have efficient algorithms that can deal with such problems that occur in such highly complex applications. In contrast to many algorithmic approaches studied in the traditional theoretical computer science literature, evolutionary algorithms have the ability to solve such complex problems [4, 38, 17, 39, 40]. Another main advantage of evolutionary algorithms is that they are easy to parallelize [36] and viewing them in the light of a tremendous number of increasing processors on multi-core computers, one can expect that the number of applications to interesting real-world problems will get a further boost during the next decade.

Viewing evolutionary algorithms from the classical theoretical computer science perspective, their main disadvantages is that their theoretical understanding lags far behind their practical success. However, in recent years, a lot of progress has been made in understanding evolutionary algorithms from a rigorous viewpoint (see the books [2, 29] for the current state of the art). With this paper we contribute to this line of research.

One of the important issues when designing successful evolutionary algorithms is to choose a suitable representation of possible solutions together with good variation operators. For problems from combinatorial optimization, different representations and variation operators have been discussed for a wide range of combinatorial optimization problems (see e.g. [30, 20, 31]). Often, variation operators (such as crossover or mutation) are designed to produce feasible offspring. For mutation this is easy to achieve, as a mutation operator usually only applies a small number of local changes to a given feasible solution.

However, the design of crossover operators, producing from two feasible solutions a new feasible one, is usually more complicated (see e.g. [24] for different crossover operators for the traveling salesman problem). Whenever a crossover operator produces an infeasible solution, one option is to discard it. However, this typically does not lead to efficient methods, as time is wasted on producing infeasible solutions and evaluating them. To deal with this situation, one can use repair mechanisms, which produce from an infeasible solution a feasible one based on properties of both parents [41]. Another way of dealing with the problem of infeasible solutions is to use specific selection methods and/or more problem specific crossover operators that are likely to produce promising solutions [5, 25].

The goal of this paper is to point out the effect of repair mechanisms and parent selection for crossover on the runtime of evolutionary algorithms in combinatorial optimization. Analyzing the runtime behavior of evolutionary algorithms has become a major part in their theoretical analysis. Based on results for different kinds of pseudo-Boolean functions [12, 18], results have been obtained for different kinds of combinatorial optimization problems [29]. Starting with some results for classical combinatorial optimization problems that are solvable in polynomial time such as the computation of minimum spanning trees [28] or maximum matchings [16], different results have been obtained for NP-hard problems [27, 15, 22, 42]. Analyses discussing the use of crossover operators on problems with a bit string representation include [19, 21]. A general recent study on the structure of crossover-based search is given in [26].

One cannot expect to beat the best known algorithms if the problem under consideration can be solved in polynomial time. With such studies we want to gain new insights on how evolutionary algorithms behave on natural optimization problems and identify into the important modules that make such algorithms successful.

Here, we carry out theoretical studies on evolutionary algorithms for the computation of shortest paths. Computing shortest paths is one of the basic problems in computer science and has already been considered in various theoretical studies of evolutionary algorithms. There are different results for the single-source shortest path (SSSP) problem [3, 32, 8].

We investigate the all-pairs shortest path (APSP) problem which is a generalization of the SSSP problem. We are given a strongly connected[1] directed graph $G = (V, E)$ with $|V| = n$ and $|E| = m$ and a weight function $w : E \to \mathbb{R}$ that assigns weights to the edges. (We distinguish between the *weight* of a path given by the sum of the weight of all its edges, and its *length*, defined as the *number* of edges in the path.) The task is to compute from each vertex $u \in V$ a weight-shortest path to every other vertex $v \in V \setminus \{u\}$. Throughout this paper, we assume that $G$ does not contain cycles of negative weight, that is, the weight function on the edges is *conservative*. The APSP problem can be solved by the Floyd-Warshall algorithm; using appropriate data structures, APSP can be computed in time $O(nm + n^2 \log n)$ (see, e.g. [23]). Our goal is not to show that evolutionary algorithms perform better than the best problem specific algorithms for this problem. Our aim is to study how general purpose algorithms can deal with the APSP problem. In particular, we want to examine the usefulness of crossover operators in evolutionary computation. By this, we want to further increase the theoretical understanding of crossover and point out how slightly different crossover operators change the runtime behavior of evolutionary algorithms for the APSP problem.

We take the APSP problem as a prominent example to show in a rigorous way how different crossover operators influence the runtime of evolutionary algorithms (measured as the number of fitness evaluations). Recently, it has been shown that the use of crossover operators provably leads to better evolutionary algorithms than evolutionary algorithms that are just based on mutation [9, 11]: The runtime for the mutation-and-crossover approach is $\Theta(n^{3.25}(\log n)^{0.25})$, which is better than the expected optimization time of $\Theta(n^4)$ of the algorithm just using mutation. In addition, [35] studied the runtime behavior of ant colony optimization for this problem and proved an upper bound of $O(n^3 \log n)$.

We will see that the evolutionary approach examined in this paper solves the APSP problem in expected optimization time $O(n^3 \log n)$, which equals the best known bound for general purpose algorithms based on ant colony optimization mentioned just above.

In Section 2 we introduce the algorithm that is subject to our analyses. We obtain a total of three variants of this algorithm by considering different crossover operators:

---

[1] Strongly connected means that there exists a directed path between any pair of vertices.

the crossover from [9, 11] is described in Section 2; crossover with repair is described in Section 3; finally, we describe crossover with parent selection in Section 4.

In the remaining sections we prove runtime bounds for our two new variants of crossover for the APSP problem. In particular, we show in Section 5.2 that our repair mechanisms speed up the optimization process to $O(n^{3.2}(\log n)^{0.2})$. Furthermore, we show 5.1 that our parent selection method leads to an optimization time of $O(n^3 \log n)$. In Section 6 we discuss the general structure of the search space that is used in deriving our runtime bounds and apply our findings to the *all-pairs bottleneck paths* problem. Finally, we conclude in Section 7.

This paper is an extension of the conference publication [10]. Here, we present full proofs and additionally discuss our structural insights in the new discussion section (Section 6).

## 2 Algorithms

For the APSP problem we examine the population-based approach introduced in [7], where each individual in the population is a path. Our goal is to evolve an initial population (a set of paths) into a population which contains, for each pair of vertices $(u, v)$ with $u \neq v$, a shortest path from $u$ to $v$ (w. r. t. to the sum of the weights of the edges). Consequently, we measure the fitness of an individual as the weight of the edges that belong to the path.

Our algorithm, called Steady State GA$_{\text{APSP}}$ (see Algorithm 1), starts with a population $\mathcal{P} := \{P_{u,v} = (u,v) | (u,v) \in E\}$ of size $|E|$, containing all paths corresponding to the edges of the given graph $G$. The variation operators produce, in each iteration, one single offspring.

The Steady State GA$_{\text{APSP}}$ decides in each iteration whether the offspring is produced by crossover or mutation dependent on a parameter $p_c$ of the algorithm. With probability $p_c$ a crossover operator is applied to two randomly chosen individuals of $\mathcal{P}$ or otherwise (with probability $1 - p_c$) mutation is used to produce the offspring. To make sure that both operators, mutation and crossover, are used we require $p_c \notin \{0, 1\}$. For all investigations in this paper, we assume that $p_c$ is chosen as an arbitrary constant, i. e. $p_c \in ]0, 1[$.

The mutation operator takes an individual $P_{x,y}$ from the population and applies sequentially $S + 1$ local operations. Here, $S$ is randomly chosen according to the Poisson distribution with parameter $\lambda = 1$. In a local operation, the current path is either lengthened or shortened by a single edge. Assume that the current individual represents a path $P_{x,y} = (x = v_0, v_1, \ldots v_{\ell-1}, y = v_\ell)$ from $x$ to $y$ consisting of $\ell$ edges, and denote by $E^-(v)$ and $E^+(v)$ the set of incoming and outgoing edges of a vertex $v$ in $G$, respectively. Then an edge $e = (u, v) \in E^-(x) \cup E^+(y) \cup \{(x, v_1), (v_{\ell-1}, y)\}$ is chosen uniformly at random. If $e \in \{(x, v_1), (v_{\ell-1}, y)\}$, the edge is removed. This means that either the first edge or the last edge in the path is removed leading to an individual $P'_{v_1,y}$ or $P'_{x,v_{\ell-1}}$ consisting of $\ell - 1$ edges. If $e \in (E^-(x) \cup E^+(y)) \setminus \{(x, v_1), (v_{\ell-1}, y)\}$, the edge is added and the path is lengthened. Here, a new individual $P'_{u,y}$ or $P'_{x,v}$ is produced that contains $\ell + 1$ edges. Note that a local operation applied to a valid path always leads to a new valid solution which implies that the mutation operator only constructs solutions which are paths.

```
1  𝒫 = {P_{u,v} = (u,v) | (u,v) ∈ E};
2  while true do
3  │   Choose r ∈ [0,1] uniformly at random;
4  │   if r ≤ p_c then
5  │   │   choose two individuals P_{x,y} and P_{x',y'} from 𝒫 u. a. r. ;
6  │   │   perform crossover on P_{x,y} and P_{x',y'} to obtain an individual P'_{s,t} ;
7  │   else
8  │   │   choose one individual P_{x,y} uniformly at random from 𝒫 and mutate P_{x,y} to
   │       obtain an individual P'_{s,t};
9  │   if P'_{s,t} is a path from s to t then
10 │   │   if there is no individual P_{s,t} ∈ 𝒫 then 𝒫 = 𝒫 ∪ {P'_{s,t}};
11 │   │   else if w(P'_{s,t}) ≤ w(P_{s,t}) then 𝒫 = (𝒫 ∪ {P'_{s,t}}) \ {P_{s,t}};
```

**Algorithm 1:** Steady State GA$_{\text{APSP}}$

Typically, crossover takes two individuals and combines them into a valid path if the end vertex of $P_{x,y}$ and the start vertex of $P_{x',y'}$ match. Choosing both individuals uniformly at random from $\mathcal{P}$, as it was done in [7, 11], often does not lead to a recombined offspring that represents a path in the given graph. In the next section, we discuss how repair mechanisms can lead to more efficient evolutionary algorithms. Later on, in Section 4, we discuss how selection methods that select promising pairs of individuals for crossover let us prove even better runtime bounds.

The selection operator only accepts individuals that are paths in the graph. In addition, it ensures diversity with respect to the different pairs of vertices. For this reason, each individual $P_{u,v}$ is indexed by the start vertex $u$ and the end vertex $v$. In the selection step an offspring is only compared to an individual of the current population that has the same start and end vertex. It is ensured that, for each pair of vertices $(u,v)$ with $u \neq v$, at most one individual $P_{u,v}$ is contained in the population. This implies that the population size of our algorithms is always at most $n(n-1)$.

For our theoretical investigations, we measure the optimization time of the algorithm by the number of fitness evaluations until an optimal population is reached for the first time. A population is optimal if it contains a a shortest path for each pair of vertices. Finally, the term w. h. p. (with high probability) denotes a result that holds with probability at least $1 - O(n^{-c})$ for some $c > 0$ independent of $n$.

## 3  Crossover with Repair

In this section, we present a simple way to increase the success probability of the crossover operator used in previous work. This results, as we shall prove rigorously in Section 5.2, is an optimization time of $O(n^{3.2}(\log n)^{0.2})$.

The main reason why previous crossover operators for the APSP problem have a rela-

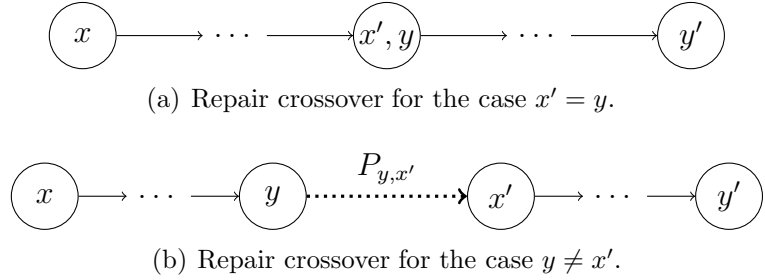

(a) Repair crossover for the case $x' = y$.

(b) Repair crossover for the case $y \neq x'$.

Figure 1: Effect of the repair crossover applied to two paths $P_{x,y}$ and $P_{x',y'}$ for the two possible cases of Algorithm 2.

tively small success probability is the fact that very often the two parent individuals simply do not fit together. That is, the end-point of the first is not equal to the starting point of the second path. Since this is a rather obvious way of failing, one might think of simple solutions.

One natural way is the following. If the end-point of the first and the starting point of the second path are different, we try to bridge this gap by the (if existent, unique) path from one point to the other which is contained in our population. If the population does not contain such a bridging path, then the crossover operator still fails. This is what we shall call *crossover with repair*.

Let the crossover operator with repair be defined as follows (see Figure 1 for a depiction of the application of the crossover operator).

**Input**: $P_{x,y} = (x, \ldots, y)$ and $P_{x',y'} = (x', \ldots, y')$
1 **if** $y = x'$ **then**
2 $\quad P'_{s,t} = (s = x, \ldots, y = x', \ldots, t = y')$ merging $P_{x,y}$ and $P_{x',y'}$ at $y$ ;
3 **else**
4 $\quad$ **if** *there is a path $P_{y,x'}$ from $y$ to $x'$ in $\mathcal{P}$* **then**
5 $\quad\quad P'_{s,t} = (s = x, \ldots, y, \ldots, x', \ldots, t = y')$ merging $P_{x,y}$, $P_{y,x'}$ and $P_{x',y'}$ at their common endpoints;
6 $\quad$ **else**
7 $\quad\quad$ the operator fails and returns a dummy individual with fitness worse than all other possible individuals;

**Algorithm 2:** Crossover with Repair.

The individual $P_{y,x'}$ from Line 4 is called *repair-path*. We refer to this operator as the *crossover with repair* or repair-crossover.

Note that this operation replaces Lines 5 in Algorithm 1.

For the variant of the Steady State $\text{GA}_{\text{APSP}}$ using the repair-operator, we obtain the following theorem.

**Theorem 1** (Crossover with Repair). *For every crossover rate $p_c \in (0,1)$ there exists a positive absolute constant $C := C(p_c)$ such that the Steady State $GA_{APSP}$ using crossover with repair (Algorithm 2) has an optimization time of at most $Cn^{3.2}(\log n)^{0.2}$ w. h. p.*

A proof of this theorem can be found in Section 5.2. As mentioned in the introduction, this result shows that the simple repair mechanism improves on the runtime of the Steady State GA$_{\text{APSP}}$ without repair (which had a runtime of $\Theta(n^{3.25}(\log n)^{0.25})$).

## 4 Feasible Parent Selection

The previous section introduced a simple repair mechanism that leads to an optimization time of $O(n^{3.2}(\log n)^{0.2})$, which is already an improvement over the optimization time of $\Theta(n^{3.25}(\log n)^{0.25})$ for the Steady State GA$_{\text{APSP}}$ in [11]. Nevertheless, the crossover operator may still produce solutions that do not constitute paths. This is the case if the start vertex of the second individual does not match the end vertex of the first individual and there is no individual in $\mathcal{P}$ for repair.

In the following, we want to make sure that the crossover operator constructs feasible solutions, i. e. individuals that represent paths. This is done by restricting the parent selection for crossover to individuals that match with respect to their endpoints. We choose the two individuals for crossover in Line 5 of Algorithm 1 using the feasible parent selection procedure given in Algorithm 3.

**1** Choose $P_{x,y} \in \mathcal{P}$ uniformly at random.
**2** Choose $P_{x',y'} \in \{P_{u,v} \mid P_{u,v} \in \mathcal{P} \land u = y \land v \neq x\}$ uniformly at random.

**Algorithm 3:** Feasible Parent Selection.

It chooses the first individual $P_{x,y}$ uniformly at random from the population $\mathcal{P}$ and the second individual $P_{x',y'}$ uniformly at random among all individuals in $\mathcal{P}$ whose start vertex equals the end vertex $y$ of $P_{x,y}$ but whose end vertex does not equal the start vertex of $P_{x,y}$. Afterwards, in Line 6 of Algorithm 1, crossover is performed by concatenation (compare the result of this operation in Figure 1(a)). Note that, due to the selection of the two individuals, a path from $x$ to $y'$ is constructed, which implies that the crossover operator only constructs feasible solutions.

This selection operator for the two parents lets us prove even better bounds on the optimization time than for crossover with repair as follows.

**Theorem 2** (Feasible Parent Selection). *For every crossover rate $p_c \in (0, 1)$ there exists a positive absolute constant $C := C(p_c)$ such that the Steady State $GA_{APSP}$ using feasible parent selection (Algorithm 3) has an optimization time of at most $Cn^3 \log n$ w. h. p.*

We will prove this theorem in Section 5.2. Recall that the bound given equals the best known bound for a randomized search heuristic on the APSP problem.

## 5 Runtime Analysis

In this section we introduce notations and mathematical methods which we will use for the proofs of Theorem 1 and Theorem 2 in the following two sections. It turns out, that in both proofs the analysis follows a common scheme. At the center of the analysis is a stage-wise analysis of the optimization process of the Steady State $\text{GA}_{\text{APSP}}$. That is, we show that the Steady State $\text{GA}_{\text{APSP}}$ passes through certain stages during its execution until it eventually reaches an stage where the population contains only shortest paths. Note that these stages are merely a concept to facilitate the runtime analysis of the Steady State $\text{GA}_{\text{APSP}}$ and do not explicitly occur in the definition (Algorithm 1) of the Steady State $\text{GA}_{\text{APSP}}$.

As in the previous sections, we again assume that we are given a a strongly connected directed graph $G = (V, E)$ on $n$ vertices and a conservative weight function $w \colon E \mapsto \mathbb{R}$, i. e. $G$ does not contain cycles of negative weight. Recall that the *weight* of a path is the sum of the weights of all its edges while its *length* is simply the number of its edges. To avoid confusion, we refer to (weight-)shortest paths in $G$ as *weight-minimal*, that is, a $u$-$v$-path $P$ is weight-minimal in $G$ if all $u$-$v$-paths in $G$ have at least the same weight as $P$.

It turns out that throughout the proofs in this and the following sections we will never regard the actual weight of a path in $G$ (although, of course, we will be constantly concerned with the presence of certain weight-minimal paths in the population). Instead, we repeatedly make use of the following observation.

**Lemma 3.** *Let $G = (V, E)$ be a finite, strongly connected directed graph with a conservative, linear weight function $w \colon E \to \mathbb{R}$. Assume that $P_{u,x}$ and $P_{x,v}$ as well as $P'_{u,x}$ and $P'_{x,v}$ are paths in $G$, and $\circ$ denotes the concatenation of two paths. If $w(P'_{u,x}) \leq w(P_{u,x})$ and $w(P'_{x,v}) \leq w(P_{x,v})$ holds, then we can deduce $w(P'_{u,x} \circ P'_{x,v}) \leq w(P_{u,x} \circ P_{x,v})$.*

*Proof.* For our linear weight function $w$ given by the definition of the APSP we automatically get

$$w(P'_{u,x} \circ P'_{x,v}) = w(P'_{u,x}) + w(P'_{x,v}) \leq w(P_{u,x}) + w(P_{x,v}) = w(P_{u,x} \circ P_{x,v}).$$

□

**Corollary 4.** *If the concatenation of two paths $P_{u,x} \circ P_{x,v}$ creates a weight-optimal path $P_{u,v}$, and if the two paths $P'_{u,x}$ and $P'_{x,v}$ are weight-optimal, too, then $P'_{u,x} \circ P'_{x,v}$ creates a weight-optimal path $P'_{u,v}$.*

*Proof.* We have $w(P'_{u,x} \circ P'_{x,v}) \leq w(P_{u,x} \circ P_{x,v})$ due to Lemma 3 and also $w(P_{u,x} \circ P_{x,v}) \leq w(P'_{u,x} \circ P'_{x,v})$ due to the optimality of $P_{u,x} \circ P_{x,v}$. □

Note, that by design our crossover operators concatenates several paths to derive $P_{u,v}$. Consequently, there is an order in which Lemma 3 and Corollary 4 can be applied such that the property of weight-minimality holds if our crossover operators are applied to weight-minimal paths. Regarding mutation we ignore the ability to shrink paths by deleting edges at the beginning or end in our analyses. Thus, the application of mutation can also be regarded as a concatenation of two suitable paths.

In order to define the different stages properly (which we do separately in the following two sections for each of the two version of the Steady State $\text{GA}_{\text{APSP}}$), we distinguish all pairs of vertices in $G$ for which there exists a weight-minimal path of given length.

**Definition 5** (Vertex Pairs $V_a^2$). *For $a \in \mathbb{R}^+$, let $V_a^2$ be the set of all pairs $(u,v) \in V^2$ with $v \neq w$ such that among all weight-minimal u-v-paths in $G$ there exists a path of length at most $a$.*

Note that in the previous definition the number $a$ can take any (positive) *real* value although the lengths of paths in $G$ is are always *integral*. We chose this definition for the sole reason of producing formally correct proofs which avoid tedious rounding operators. For the understanding of these proofs, however, we may think of $a$ as being integral. In particular, we have that $V_a^2 = V_{\lfloor a \rfloor}^2$ for all $a \in \mathbb{R}^+$. Moreover, since $G$ is strongly connected and a path in $G$ is at most of length $n-1$, it holds that

$$V_a^2 = \{(u,v) \in V^2 : v \neq w\}. \tag{1}$$

for all $a \in \mathbb{R}^+$ with $a \geq n-1$.

At this point, let us also introduce a probabilistic tool we will repeatedly use in the proof of Theorem 1 and Theorem 2. The following Lemma is an adaptation of the Coupon Collector argument. It allows us to give a lower bound on the probability of hitting each element of a set $I$ using $r$ samples, provided the probability to sample an element that has not yet been sampled is bounded below by a positive constant.

**Lemma 6** (Coupon Collector with Dependencies).
*Let $I$ be a finite set, $p \in (0,1)$, and $\{A_i^t\}_{t \in \mathbb{N}}$ be sequences of events indexed by $I$ such that*

$$\Pr\left[A_i^t \,\Big|\, \bigcap_{0 \leq s < t} \overline{A_i^s}\right] \geq p$$

*holds for all $t \in \mathbb{N}$ and all $i \in I$.*

*If $T$ is the random variable that denotes the first point in time $t \in \mathbb{N}$ such that for all $i \in I$ one of the events $A_i^0, \ldots, A_i^t$ has occurred, then it holds for all $r \in \mathbb{R}^+$ that*

$$\Pr[T \geq r] \leq |I| \cdot \mathrm{e}^{-pr}.$$

*Proof.* First, we show the lemma for $r \in \mathbb{N}$. For each $i \in I$, let $B_i$ be the event that none of the events $A_i^0, \ldots, A_i^{r-1}$ occurs. Then

$$\Pr[B_i] = \Pr\left[\bigcap_{0 \leq t < r} \overline{A_i^t}\right] = \prod_{0 \leq t < r} \Pr\left[\overline{A_i^t} \,\Big|\, \bigcap_{0 \leq s < t} \overline{A_i^s}\right].$$

holds for all $i \in I$ and thus

$$\Pr[B_i] \leq (1-p)^r \leq \mathrm{e}^{-pr}.$$

Since the two events "$T \geq r$" and "$\bigcup_{i \in I} B_i$" coincide, we obtain by the Union Bound (see [1]) that

$$\Pr[T \geq r] \leq \sum_{i \in I} \Pr[B_i] \leq |I| \cdot \mathrm{e}^{-pr}.$$

Now, the lemma also follows for arbitrary positive real values of $r$,

$$\Pr[T \geq r] = \Pr[T \geq \lceil r \rceil] \leq |I| \cdot \mathrm{e}^{-p\lceil r \rceil} \leq |I| \cdot \mathrm{e}^{-pr}$$

holds for all $r \in \mathbb{R}^+$. □

In the following two sections, we present the proofs of Theorem 1 and Theorem 2, that is, we give upper bounds on the optimization times of the Steady State $\text{GA}_{\text{APSP}}$ using the operator "Crossover with Repair" and the operator "Feasible Parent Selection", respectively. Since the proof of Theorem 1 is more involved, we start with the proof of Theorem 2 in the next section.

## 5.1 The Proof of the Runtime Bound for Feasible Parent Selection

This section is devoted to the proof Theorem 2. For simplicity, whenever we refer to the Steady State $\text{GA}_{\text{APSP}}$ in this section, we assume without further mentioning that the operator "Feasible Parent Selection" is applied. As discussed above, we want to analyze the behavior of this algorithm in stages. To this end, we say the Steady State $\text{GA}_{\text{APSP}}$ has completed the $k$-th stage if the population contains a weight-minimal $u$-$v$-path for every pair of vertices $(u, v)$ for which there exists such a weight-minimal path of length at most $a(k)$ in the graph $G$, where the sequence $\{a(k)\}_{k \in \mathbb{N}}$ is given by

$$a(k) := (3/2)^k \tag{2}$$

for all $k \in \mathbb{N}$. Clearly, the first point in time when this event happens defines a random variable. We call this random variable $T_k$ and say it marks the end of the $k$-th stage. The following definition makes this notion precise.

**Definition 7** (Time $T_k$). *For $k \in \mathbb{N}$, let $a(k)$ be as defined in (2) and let $T_k$ be the random variable that denotes the first point in time such that, for all pairs $(u, v) \in V^2_{a(k)}$, the population of the Steady State $GA_{APSP}$ contains a weight-minimal $u$-$v$-path.*

Observe crucially that, although $(u, v) \in V^2_{a(k)}$ implies there exists a weight-minimal path of length at most $a(k)$ in $G$, we only require the existence of any weight-minimal $u$-$v$-path in the population of the Steady State $\text{GA}_{\text{APSP}}$ at time $T_k$. In particular, this $u$-$v$-path may be arbitrarily long.

It is clear by the definition of the $V^2_{a(k)}$'s (Definition 5) that

$$T_0 \leq T_1 \leq T_2 \leq \ldots$$

holds. Also note that, depending on $G$, the sets $V^2_{a(k)}$ and $V^2_{a(k+1)}$ may be equal for some values of $k$. In this case, also the random variables $T_k$ and $T_{k+1}$ coincide, that is, we have $T_k = T_{k+1}$.

We have already seen in (1) that if $a(k) \geq n - 1$, then the population of the Steady State $\text{GA}_{\text{APSP}}$ contains only weight-minimal paths. Since $a(n) \geq n - 1$ for all values of $n$, this implies that at time $T_n$ latest, the Steady State GA$_{\text{APSP}}$ has found a population of weight-minimal paths. In other words, the random variable $T_n$ dominates the optimization time of the Steady State GA$_{\text{APSP}}$. Thus, in order to show Theorem 2, it is sufficient to show that, for every $p_c \in (0,1)$, there exists an positive absolute constant $C := C(p_c)$ such that

$$\Pr\left[T_n \geq Cn^3 \log n\right] \leq \frac{1}{n}. \tag{3}$$

In fact, we show an even stronger statement, given in the following proposition.

**Proposition 8.** *For every $p_c \in (0,1)$, there exists an positive absolute constant $C_1 := C_1(p_c)$ such that,*

$$\Pr\left[T_k \geq T_{k-1} + C_1(2/3)^k n^3 \log n + 1\right] \leq \frac{1}{n^2}. \tag{4}$$

*holds for all $k \in \{1, \ldots, n\}$.*

Before we prove Proposition 8, let us first argue how it implies (3) and thus Theorem 2. To this end, consider the telescopic sum

$$T_n = T_0 + \sum_{k=1}^{n} \left(T_k - T_{k-1}\right).$$

The random variable $T_0$ denotes the first point in time such that the population contains a weight-minimal path for all pairs $(u,v)$ which have a weight-minimal path of length $a(0) = 1$. But such a path of length 1 consists only of a single (directed) edge, and for all vertex pairs that form such an edge this edge is present in the initial population (Step 1 in Algorithm 1). Therefore, we always have $T_0 = 0$. Next, suppose that

$$T_k - T_{k-1} \leq C_1(2/3)^k n^3 \log n + 1$$

holds for all $k \in \{1, \ldots, n\}$. In this case, we have

$$T_n \leq C_1\Big(\sum_{k=1}^{n}(2/3)^k\Big)n^3 \log n + n \leq 2n^3 \log n + n,$$

where we bounded $\sum_{k=1}^{n}(2/3)^k$ by the geometric sum

$$\sum_{k=1}^{\infty}(2/3)^k \leq 2.$$

Thus, if we set $C := 2C_1 + 1$, then the event

"$\forall k \in \{1, \ldots, n\}\colon T_k - T_{k-1} \leq C_1(2/3)^k n^3 \log n + 1$"

implies the event "$T_n \leq Cn^3 \log n$". Conversely, this means that the event "$T_n \geq Cn^3 \log n$" implies the event

$$\text{"} \exists k \in \{1, \ldots, n\} \colon T_k - T_{k-1} \geq C_1(2/3)^k n^3 \log n + 1 \text{"}.$$

Therefore, we have

$$\Pr\left[T_n \geq Cn^3 \log n\right] \leq \Pr\left[\exists k \in \{1, \ldots, n\} \colon T_k - T_{k-1} \geq 1 + C_1(2/3)^k n^3 \log n\right]$$

and (3) follows from Proposition 8 by the Union Bound (see [1]).

In order to conclude the proof of Theorem 2, we still need to show Proposition 8. We devote the remainder of this section to this proof.

Proposition 8 basically states that with sufficiently high probability, the time needed by the Steady State $\text{GA}_{\text{APSP}}$ between the $(k-1)$-th stage and the $k$-th stage is not too long. Interestingly, for the version of the Steady State $\text{GA}_{\text{APSP}}$ we are considering in this section (the one with feasible parent selection), it is sufficient to regard only the effect of crossover and to neglect the effect of mutation (which can only be beneficial to the analysis) on the optimization process.

In our analysis, we strongly rely on the following considerations. If we perform (feasible) parent selection and subsequent crossover on a population which contains all weight-minimal paths of length at most $a(k-1)$, then it is sufficiently likely to generate each weight-minimal path of length at most $a(k) = (3/2)a(k-1)$. Now, the crucial observation is that this argument remains valid if the population contains weight-minimal paths of *any* length for every pair of vertices that has a weight-minimal paths of length at most $a(k-1)$ in the graph $G$. Of course, in this case we cannot guarantee the Steady State $\text{GA}_{\text{APSP}}$ produces every weight-minimal path of length at most $a(k)$. However it will instead produce some weight-minimal path for every pair of vertices that has a weight-minimal paths of length at most $a(k)$ in the graph $G$. This observation is formalized in the following lemma.

**Lemma 9.** *For every $a \in \mathbb{R}^+$ with $a \geq 1$ and every pair $(u, v) \in V^2_{(3/2)a} \setminus V^2_a$, there exists at least $a/4$ different vertices $x \in V$ such that $(u, x)$ and $(x, v)$ are in $V^2_a$ and such that the concatenation of every weight-minimal u-x-path with every weight-minimal x-v-path at the vertex $x$ results in a weight-minimal u-v-path.*

*Proof.* Let the vertex pair $(u, v)$ be in the set $V^2_{(3/2)a}$ but not in the set $V^2_a$. By the definitions of $V^2_a$ and $V^2_{(3/2)a}$, there exists a weight-minimal $u$-$v$-path $P_{u,v} = (u = u_0, u_1 \ldots, u_\ell = v)$ of length $\ell$ with

$$a < \ell \leq (3/2)a.$$

Consider the index set

$$J := \{\lceil a/2 \rceil, \ldots, \lfloor a \rfloor\}.$$

Let $j \in J$ and $x := u_j$. Note that $\lfloor a \rfloor \leq a < \ell$ and therefore $u_j$ is a well-defined vertex of $P_{u,v}$. By Corollary 4, the two paths $P_{u,x} = (u = u_0, \ldots, u_j = x)$ and $P_{x,v}(x = u_j, \ldots, u_\ell = v)$ are weight-minimal, too. Moreover, both paths are of length at most $a$ since $j \leq a$ and also $\ell - j \leq (3/2)a - a/2 = a$. Hence, $(u,x)$ and $(x,v)$ are in $V_a^2$. Furthermore, again by Lemma 3, the concatenation of any two weight-minimal paths $P'_{u,x}$ and $P'_{x,v}$ (of arbitrary length) at the vertex $x$ is weight-minimal, too. Now, all that is left to do is to bound the number of possible choices for $x$. Since

$$|J| = \lfloor a \rfloor - \lceil a/2 \rceil + 1 \geq \begin{cases} 1 - 1 + 1 \geq a/4 & \text{if } 1 \leq a < 2, \\ 2 - 2 + 1 \geq a/4 & \text{if } 2 \leq a < 4, \\ a - 1 - (a/2) - 1 + 1 \geq a/4 & \text{if } a \geq 4, \end{cases}$$

there are at least $a/4$ ways to choose $x$, which concludes the proof of this lemma. □

Finally, with Lemma 9 at hand, we are ready to prove Proposition 8 which will also conclude the proof of Theorem 2.

*of Proposition 8.* Let $k \in \{1, \ldots, n\}$. We show that (4) holds, that is, we show that there exists a positive absolute constant $C_1 := C_1(p_c)$ (to be chosen later) such that the event

$$\text{``}T_k \geq T_{k-1} + 1 + C_1(2/3)^k n^3 \log n\text{''} \tag{5}$$

occurs with probability at most $1/n^2$.

At time $T_{k-1}$, which marks the beginning of the $k$-th stage, we have for every pair of vertices in $V_{a(k-1)}^2$ a weight-minimal path in the population of the Steady State GA$_{\text{APSP}}$. Now, we want to bound the duration of the $k$-th stage, that is, number of iterations needed until the population also contains a weight-minimal path for every pair of vertices in $V_{a(k)}^2$.

As a consequence Lemma 9, we will see below that in each iteration of the $k$-th stage the probability that the Steady State GA$_{\text{APSP}}$ produces a weight-minimal path for a pair in $V_{a(k)}^2 \setminus V_{a(k-1)}^2$ is at least $(1/6)(3/2)^k p_c n^{-3}$. Thus, informally speaking, we would expect a kind of Coupon Collector process to happen, which produces all such weight-minimal paths (there are at most $n^2$) in an expected number of $6p_c^{-1}(2/3)^k n^3 \log n^2$ iterations.

However, since we only have lower bounds on the probabilities to find a pair in $V_{a(k)}^2 \setminus V_{a(k-1)}^2$ and since the events we regard are not independent of each other, we have to be more careful in bounding the probability of the event that (5) holds. For this reason, we apply Lemma 6.

In the notation of Lemma 6, let $I = V_{a(k-1)}^2 \setminus V_{a(k)}^2$ and let $A_{(u,v)}^t$ with $t \in \mathbb{N}$ and $(u,v) \in I$ denote the event that at time $T_{k-1} + 1 + t$ there exists a weight-minimal $u$-$v$-path in the population of the Steady State GA$_{\text{APSP}}$. Then, we show that there exists a $p \in (0,1)$ such that

$$\Pr\left[A_{(u,v)}^t) \,\bigg|\, \bigcap_{0 \leq s < t} \overline{A_{(u,v)}^s}\right] \geq p \tag{6}$$

holds for all $t \in \mathbb{N}$ and $(u,v) \in I$.

For this, first note that since $(u,v)$ is not in $V_{a(k)}^2$, with positive probability there is no weight-minimal $u$-$v$-path in the population of the Steady State GA$_{\text{APSP}}$ at time $T_{k-1} + t$. Therefore, the conditional probability above is well-defined.

Next, since there exists a weight minimal-path for every vertex pair in $V_{a(k-1)}^2$, Lemma 9 gives us that there are at least $a(k-1)/4$ distinct vertices $x \in V$ such that $(u,x)$ and $(x,v)$ are in $V^2_{a(k-1)}$ and the concatenation of a weight-minimal $u$-$x$-path with a weight-minimal $x$-$v$-path yields a weight-minimal $u$-$v$-path. For each of these vertices $x$, the probability that the Steady State $\text{GA}_{\text{APSP}}$ performs this particular concatenation at time $T_{k-1}+1+t$ is the probability to (i) perform a crossover step, (ii) choose the weight-minimal $u$-$x$-path as first parent, and (iii) then the weight-minimal $x$-$v$-path as second parent from the population at time $T_{k-1}+t$. The event (i) has probability $p_c$. By the definition of the operator "Feasible Parent Selection" (Algorithm 3) event (ii) happens with probability at least $1/n^2$ and event (iii) with probability at least $1/n$. Moreover, these bounds are independent of the event whether or not there already exists a weight-minimal $u$-$v$-path in the population at time $T_{k-1}+t$. Thus, Equation (6) indeed holds for all $(u,v) \in I$ and $t \in \mathbb{N}$ with

$$p := \frac{a(k-1)p_c}{4n^3} = \frac{(3/2)^k p_c}{6n^3}.$$

Finally, since the event that at time $T_{k-1}+1+t$ one of the events $A^0_{(u,v)}, \ldots, A^t_{(u,v)}$ has occurred for all $(u,v) \in I$ implies the event that $T_k \leq T_{k-1}+1+t$, we have by Lemma 6 with $r := C_1(2/3)^k n^3 \log n$ and $C_1 := \frac{24}{p_c}$ that

$$pr = \frac{(3/2)^k p_c}{6n^3} \cdot \frac{24}{p_c} \cdot (2/3)^k n^3 \log n = 4 \log n \geq 4 \ln n$$

and therefore, since $I \leq n^2$ that

$$\Pr\left[T_k \geq T_{k-1}+1+C_1(2/3)^k n^3 \log n\right] \leq |I|\mathrm{e}^{-4\ln n} \leq \frac{1}{n^2}.$$

This concludes the proof of Proposition 8 and therefore also the proof of Theorem 2. □

## 5.2 The Proof of the Runtime Bound for Crossover with Repair

We now prove Theorem 1. Again for the sake of brevity, whenever we refer to the Steady State $\text{GA}_{\text{APSP}}$ in this section, we now assume without further mentioning that the operator "Crossover with Repair" is applied. The proof will follow the same line of argument like that in the previous section. That is, we again want to analyze the behavior of this algorithm in stages. However, this time we say the Steady State $\text{GA}_{\text{APSP}}$ has completed the $k$-th stage if the population contains a weight-minimal $u$-$v$-path for every pair of vertices $(u,v)$ for which there exists such a weight-minimal path of length at most $b(k)$ in the graph $G$, where the sequence $\{b(k)\}_{k \in \mathbb{N}}$ is now given by

$$b(k) := 24(3/2)^k (n \log n)^{1/5} \tag{7}$$

for all $k \in \mathbb{N}$. Notice the extra $24(n \log n)^{1/5}$-factor in the definition above compared to the definition of $a(k)$ in (2).

Analogously to the previous section, we let the random variable $T'_k$ mark the end of the $k$-th stage. Consequently, the following definition differs from Definition 10 only in the choice of the sequence $\{b(k)\}_{k \in \mathbb{N}}$ instead of the sequence $\{a(k)\}_{k \in \mathbb{N}}$.

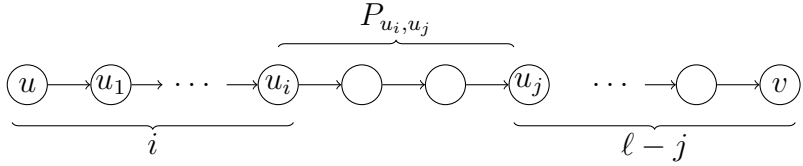

Figure 2: $P_{u_i,u_j}$ is an approximating path for the vertex pair $(u,v)$ with a gap of $g := \max\{i, \ell - j\}$.

**Definition 10** (Time $T'_k$). *For $k \in \mathbb{N}$, let $b(k)$ be as defined in (7) and let $T'_k$ be the random variable that denotes the first point in time such that, for all pairs $(u,v) \in V^2_{b(k)}$, the population of the Steady State $GA_{APSP}$ contains a weight-minimal $u$-$v$-path.*

In the previous section, we saw that the Steady State $GA_{APSP}$ applying the Feasible Parent Selection operator can find all minimum-weight shortest paths necessary to complete the current stage by crossover only. For the Steady State $GA_{APSP}$ which we consider in this section (the one applying the operator "Crossover with Repair"), our analysis is slightly more involved since it includes both, crossover and mutation.

To capture the interplay between crossover and mutation, we divide all stages into two phases, which we call the *crossover* phase and the *mutation phase*. Like before with the stages, the distinction of phases are merely a method of analysis and do not change the definition of Algorithm 1. This means that in our analysis of the optimization behavior of the Steady State $GA_{APSP}$ we will only consider the effect of crossover during the crossover phases and only the effect of mutation during the mutation phases. In the actual run of the algorithm, however, both crossover and mutation are likely to happen in all phases. Still, since both operators never remove a weight-minimal path from the population, we are save to ignore them in our analysis.

In order to define the two phases, we now define the notions of *gaps* and *approximating paths*, two concepts which were introduced in [11]. The key observation behind these notions is that it suffices that crossover finds a path that sufficiently well approximates a sought-after path, because mutation is fast enough to fill the gaps.

**Definition 11** (Approximating Path with Gap $g$). *Let $P_{u,v} = (u = u_0, u_1, \ldots, u_\ell = v)$ be a weight-minimal path in $G$ of length $\ell$. Suppose that $P_{x,y}$ is a path in $G$ with $x = u_i$ and $y = u_j$ for some indices $0 \leq i \leq j \leq \ell$. Then we call $P_{x,y}$ an* approximating *path for the pair $(u,v)$ with (integral) gap $g := \max\{i, \ell - j\}$; compare Figure 2.*

Notice crucially that it is not necessary that an approximating path for the pair $(u,v)$ approximates the *weight* of a weight-minimal $u$-$v$-path. In particular, every $u$-$v$-path is an approximating path with gap 0 for the vertex pair $(u,v)$.

With the notion of approximating paths at hand, we define the two phases of the $k$-th stage. The first phase, which we call the *crossover phase*, start with the beginning of the $k$-th stage (at time $T'_{k-1}$) and end when for all pairs $(u,v) \in V^2_{b(k)}$, there exists a weight-minimal approximating path for $(u, v)$ with gap at most

$$g(k) := (5/6)^k (n \log n)^{1/5} \tag{8}$$

for all $k \in \mathbb{N}$.

The second phase, the *mutation* phase, lasts from the end of the crossover phase to the end of the $k$-th stage. Corresponding to $T'_k$, we define the random variable $T''_k$ which marks the end of the crossover phase and the beginning of the mutation phase.

**Definition 12** (Time $T''_k$). *For $k \in \mathbb{N}$ with $k \geq 1$, let $g(k)$ be as defined in (8) and let $T''_k$ be the random variable that denotes the first point in time after $T'_{k-1}$ such that, for all pairs $(u, v) \in V^2_{b(k)}$, the population of the Steady State $GA_{APSP}$ contains a weight-minimal approximating path for the vertex pair $(u, v)$.*

It is clear by Definition 10 and Definition 12 that

$$T'_0 \leq T''_1 \leq T'_1 \leq T''_2 \leq T'_2 \leq \ldots$$

holds and that some of these inequalities may happen to be tight. Like in the previous section, it still holds that at time $T'_n$, the Steady State $GA_{APSP}$ has found a population of weight-minimal paths. This time, however, we bound the number of phases more careful. Let $k^* := k^*(n)$ be the minimum integer $k$ such that $b(k) \geq n - 1$. Then already at time $T'_{k^*}$ the Steady State $GA_{APSP}$ has found a population of weight-minimal paths. It is easy to see that

$$k^* \leq 2 \log n.$$

In order to show Theorem 1, it is again sufficient to show that, for every $p_c \in (0, 1)$, there exists an positive absolute constant $C := C(p_c)$ such that

$$\Pr\left[T'_{k^*} \geq Cn^3 (n \log n)^{1/5}\right] \leq \frac{1}{n}. \tag{9}$$

Again, we show a stronger statement. The following proposition is the direct counterpart to Proposition 8 in the previous section. However, this time we regard both, crossover and mutation. Both operators have constant probability to be applied in an iteration, and neither can decrease the fitness of an individual. Hence we may occasionally only regard the effect of one of the two.

We start with the consideration of the effects of mutation only. We apply a result in [9] to arrive at a population that contains with high probability a weight-minimal path for every vertex pair for which there exists a weight-minimal path in $G$ of length at most $24(n \log n)^{1/5}$. This marks the end of the 0th-stage.

The duration of this initial stage is comparable to the duration of all next stages, during which the interplay between crossover and mutation plays a role. From this point on we consider the effects of the repair-crossover which allows for the creation of approximating weight-minimal path in the crossover phase of a stage. In the subsequent mutation phase, the mutation operator will close the gap between the approximating weight-minimal paths and the weight-minimal paths we are actually aiming for (again, we do not give a full proof for this but refer to [9] instead). The following proposition gives a precise statement of these observations for the initial mutation stage (Equation 10), the $k$th crossover phase (Equation 11) and the $k$th mutation phase (Equation 12).

**Proposition 13.** *For every $p_c \in (0,1)$, there exists three positive absolute constants $C_1 := C_1(p_c)$, $C_2 := C_2(p_c)$, and $C_3 := C_3(p_c)$ such that the three inequalities*

$$\Pr\left[T_0' \geq C_1 n^3 (n \log n)^{1/5} + C_1 n^3 \log n\right] \leq \frac{1}{n^2}, \tag{10}$$

$$\Pr\left[T_k'' - T_{k-1}' \geq C_2 (4/5)^{2k} n^3 (n \log n)^{1/5} + 1\right] \leq \frac{1}{n^2}, \tag{11}$$

$$\Pr\left[T_k' - T_k'' \geq C_3 (5/6)^k n^3 (n \log n)^{1/5} + C_3 n^3 \log n\right] \leq \frac{1}{n^2} \tag{12}$$

*hold for all $k \in \{1, \ldots, k^*\}$.*

As before, we defer the proof of Proposition 13 until we have shown how Proposition 13 implies (9) and thus Theorem 1. This follows exactly the same line of proof as in the previous section. We consider the telescopic sum

$$T_{k*}' = T_0' + \sum_{k=1}^{k^*} \left(T_k' - T_k''\right) + \sum_{k=1}^{k^*} \left(T_k'' - T_{k-1}'\right).$$

and note that this time we do not necessarily have $T_0' = 0$ due to the initial stage. Consider the event that the three inequalities

$$T_0' \leq C_1 n^3 (n \log n)^{1/5} + C_1 n^3 \log n,$$
$$T_k'' - T_{k-1}' \leq C_2 (4/5)^{2k} n^3 (n \log n)^{1/5} + 1,$$
$$T_k' - T_k'' \leq C_3 (5/6)^k n^3 (n \log n)^{1/5} + C_3 n^3 \log n$$

hold for all $k \in \{1, \ldots, k^*\}$. If this event occurs, then we have

$$\sum_{k=1}^{k^*} (T_k'' - T_{k-1}') \leq \left(\sum_{k=1}^{k^*} (4/5)^{2k}\right) C_2 n^3 (n \log n)^{1/5} + k^*,$$

$$\sum_{k=1}^{k^*} (T_k' - T_k'') \leq \left(\sum_{k=1}^{k^*} (5/6)^k\right) C_3 n^3 (n \log n)^{1/5} + C_3 k^* n^3 \log n.$$

Substituting $k^* \leq 2 \log n$ and bounding the geometric series' yields

$$T_{k*}' \leq (C_1 + (16/9)C_2 + 5C_3) n^3 (n \log n)^{1/5} + C_3 n^3 (\log n)^2 + 2 \log n$$

There exists a positive absolute constant $D$ such that

$$C_3 n^3 (\log n)^2 + 2 \log n \leq D n^3 (n \log n)^{1/5}$$

holds for all $n \in \mathbb{N}$. Then, for $C := 2 + C_1 + (16/9)C_2 + 5C_3 + D$, we apply the same union bound argument as in the previous section. Note that we consider the union of $2k^*+1$, that is, of at most $n$ events. Thus (9) follows from the three inequalities (10), (11), and (12) in Proposition 13. Since inequality (9) implies Theorem 1, we may conclude the proof of Theorem 1 by showing Proposition 13.

To derive the two inequalities (10) and (12) in Proposition 13, we apply a result from [9], which we do not prove here but simply state as Lemma 14. We adapt the notation to our needs, a closer look into the proofs in [9] shows that this is easily possible.

**Lemma 14** (Analysis of Mutation). *For all $p_c \in (0,1)$ there exists a positive absolute constant $C := C(p_c)$ such that the following statement holds.*

*Let $g \in \mathbb{R}^+$ and let $W \subseteq V_n^2$. Suppose that at time $t_0 \in \mathbb{N}$ the population of the Steady State $GA_{APSP}$ contains a weight-minimal approximating path with gap at most $g$ for every vertex pair $(u,v) \in W$. Let $T$ be the random variable that denotes the first point in time such that the population of the Steady State $GA_{APSP}$ contains a (proper) weight-minimal path for every vertex pair $(u,v) \in W$. Then*

$$\mathrm{Prob}\Big[T \geq t_0 + Cn^3(g + \log n)\Big] \leq \frac{1}{n^2}.$$

The two inequalities (10) and (12) in Proposition 13 directly follow from the previous lemma.

*of inequality (10) in Proposition 13.* Let $C_1 := 24C$ where $C := C(p_c)$ is the positive absolute constant provided by Lemma 14. Then inequality (10) follows from Lemma 14 if we set $g := b(0)$, $W := V_{b(0)}^2$, and $t_0 = 0$. □

*of inequality (12) in Proposition 13.* Let $k \in \{1, \ldots, k^*\}$ and let $C_1 := C$ where $C := C(p_c)$ is the positive absolute constant provided by Lemma 14. Then inequality (12) again follows from Lemma 14 if we set $g := g(k)$, $W := V_{b(k)}^2$, and replace $t_0$ by the random variable $T'_{k-1}$ (we may do this by the law of total probability, since Lemma 14 holds for every choice of $t_0$). □

At this point, we are only left to prove inequality (11) in Proposition 13. This is done similarly to the proof of Proposition 8 and will be the remainder of this section. In particular, we next show a lemma which is the direct counterpart to Lemma 9 in the previous section.

The following lemma gives a lower bound on the number of different combinations how the Steady State $GA_{APSP}$ can produce a weight-minimal approximating path with gap $g(k)$ for any weight-minimal path of length at most $b(k) = (3/2)b(k-1)$. Again, we will assume that for every weight-minimal graph of length at most $b(k-1)$ there exist a weight minimal path (of arbitrary length) in the population of the Steady State $GA_{APSP}$ at that time.

**Lemma 15.** *For every $b \in \mathbb{R}^+$ with $b \geq 24$, every $g \in \mathbb{R}^+$ with $g \leq b/12$, and every pair $(u,v) \in V_{(3/2)b}^2 \setminus V_b^2$, there exists at least $(bg)^2/36$ different tuples $(u', x, y, v')$ of four distinct* vertices $u', x, y, v' \in V$ such that the pairs $(u', x)$, $(x, y)$, and $(y, v')$ are in $V_b^2$ and such that the concatenations of every weight-minimal $u'$-$x$-path and every weight-minimal $y$-$v'$-path with every weight-minimal $x$-$y$-path at the vertices $x$ and $y$ results in a weight-minimal $u'$-$v'$-path which is an approximating path with gap at most $g$ for the pair $(u, v)$; compare Figure 3.

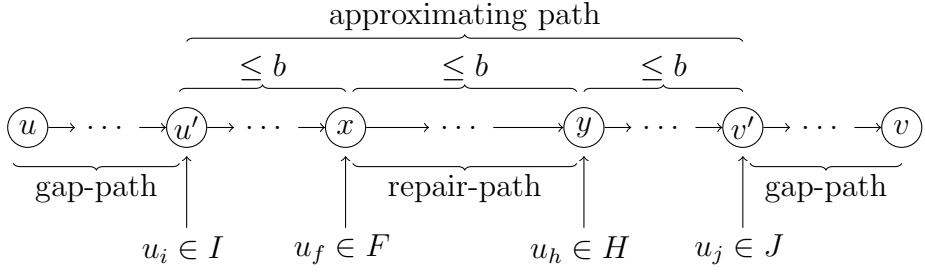

Figure 3: The path depicts the situation of Lemma 15 with two possible gap paths and a repair path from $x$ to $y$. The $x$-$y$-repair-path stems from the paths from $u'$ to $x$ and from $y$ to $v'$ which are input of the repair-crossover. Each of these three paths has length at most $b$ and their concatenation is a approximating path with gap $g$ for the vertex pair $(u, v)$.

*Proof.* Let the vertex pair $(u, v)$ be in the set $V_{(3/2)b}^2$ but not in the set $V_b^2$. By the definitions of $V_b^2$ and $V_{(3/2)b}^2$, there exists a weight-minimal $u$-$v$-path $P_{u,v} = (u = u_0, u_1 \ldots, u_\ell = v)$ of length $\ell$ with
$$b < \ell \leq (3/2)b\,.$$
Consider the four index sets
$$\begin{aligned} I &:= \{0, \ldots, \lfloor g \rfloor\}, \\ F &:= \{\lceil b/2 \rceil, \ldots, \lceil b/2 \rceil + \lceil b/6 \rceil - 1\}, \\ H &:= \{\lceil b/2 \rceil + \lceil b/6 \rceil, \ldots, \lceil b/2 \rceil + 2\lceil b/6 \rceil - 1\}, \\ J &:= \{\ell - \lfloor g \rfloor, \ldots, \ell\}. \end{aligned}$$
Since $g \leq b/12$ and $b \geq 24$, we have that
$$g \leq b/12 < (b/12) + 1 \leq \lceil b/2 \rceil,$$
and, since $\ell > b$, we have that
$$\lceil b/2 \rceil + 2\lceil b/6 \rceil - 1 \leq 5b/6 + 2 \leq b - (b/12) < \ell - \lfloor g \rfloor, \tag{13}$$
the index sets $I, F, H$, and $J$ are disjoint and subsets of $\{0, \ldots, \ell\}$. Let $i \in I$, $f \in F$, $h \in H$, and $j \in J$. Furthermore, let $u' := u_i$, $x := u_f$, $y := u_h$, and $v' := u_j$. Then, by repeated application of Lemma 3 and Corollary 4, the three paths $P_{u',x} = (u' = u_i, \ldots, u_f = x)$, $P_{x,y} = (x = u_f, \ldots, u_h = y)$ and $P_{y,v'} = (y = u_h, \ldots, u_j = v')$ are weight-minimal, too.

Moreover, all three paths are of length at most $b$ since, by (13), we have both $f \leq h \leq b$ and $h - f \leq h \leq b$. Furthermore, as in the proof of Lemma 9, we have also also $\ell - h \leq b$. Hence, $(u', x)$, $(x, y)$, and $(y, v')$ are in $V_b^2$. Furthermore, again by repeated application of Lemma 3, the concatenation of any three weight-minimal paths $P'_{u',x}$, $P'_{x,y}$ and $P'_{y,v'}$ (of arbitrary length) at the vertices $x$ and $y$ is weight-minimal, too. Finally, there are

$$|I| \cdot |F| \cdot |H| \cdot |J| = (\lfloor g \rfloor + 1)^2 \lceil b/6 \rceil^2 \geq g^2 b^2 / 36$$

ways to choose the tuple $(u', x, y, v')$, which concludes the proof of this lemma. □

Finally, all we are left to do is to prove of inequality (11) in Proposition 13. As announced above, this proof will again follow the lines of the proof of Proposition 8 in the previous section.

*of inequality (11) in Proposition 13.* Let $k \in \{1, \ldots, k^*\}$. We show that (11) holds, that is, we show that there exists a positive absolute constant $C_2 := C_2(p_c)$ (to be chosen later) such that the event

$$\text{"} T_k'' \geq T_{k-1}' + 1 + C_2 (4/5)^{2k} n^3 (n \log n)^{0.2} \text{"} \tag{14}$$

occurs with probability at most $1/n^2$.

At time $T_{k-1}'$, which marks the beginning of the $k$-th stage, we have for every pair of vertices in $V_{b(k-1)}^2$ a weight-minimal path in the population of the Steady State GA$_{\text{APSP}}$. Now, we want to bound the duration of the crossover phase of the $k$-th stage, that is, number of iterations needed until the population also contains a weight-minimal approximating path with gap at most $g(k)$ for every pair of vertices in $V_{b(k)}^2$.

We again apply Lemma 6, let $I = V_{b(k-1)}^2 \setminus V_{b(k)}^2$ and let $A_{(u,v)}^t$ with $t \in \mathbb{N}$ and $(u, v) \in I$ denote the event that at time $T_{k-1}' + 1 + t$ there exists a weight-minimal approximating path with gap at most $g(k)$ for the pair $(u, v)$ in the population of the Steady State GA$_{\text{APSP}}$. We show that there exists a $p \in (0, 1)$ such that

$$\Pr \left[ A_{(u,v)}^t \; \Big| \; \bigcap_{0 \leq s < t} \overline{A_{(u,v)}^s} \right] \geq p \tag{15}$$

holds for all $t \in \mathbb{N}$ and $(u, v) \in I$.

For this, first note that since $(u, v)$ is not in $V_{b(k)}^2$, with positive probability there is no weight-minimal $u$-$v$-path in the population of the Steady State GA$_{\text{APSP}}$ at time $T_{k-1}' + t$. Therefore, the conditional probability above is again well-defined.

Next, since there exists a weight minimal-path for every vertex pair in $V_{b(k-1)}^2$, Lemma 15 gives us that there are at least $(b(k-1)g(k))^2/36$ distinct tuples $(u', x, y, v')$ of vertices $u', x, y, v' \in V$ such that $(u', x)$, $(x, y)$, and $(x', v)$ are in $V_{b(k-1)}^2$ and the concatenation of a weight-minimal $u'$-$x$-path, a weight-minimal $x$-$y$-path, and a weight-minimal $y, v'$-path at $x$ and $y$ yields a weight-minimal $u'$-$v'$-path which approximates the pair $(u, v)$ with gap at most $g(k)$. Note that, in order to apply Lemma 15, we need here that $b(k-1) \geq 24$, which holds since

$$b(k-1) \geq b(0) = 24(n \log n)^{0.2}$$

and that $g(k) \leq b(k-1)/12$, which holds since

$$g(k) \leq g(0) \leq 24b(0) \leq b(k-1)/12.$$

For each of these vertex tuples $(u', x, y, v')$, the probability that the Steady State $\text{GA}_{\text{APSP}}$ performs this particular concatenation at time $T'_{k-1} + 1 + t$ is the probability to (i) perform a crossover step, (ii) choose the weight-minimal $u'$-$x$-path as first parent, and (iii) then the weight-minimal $y$-$v'$-path as second parent from the population at time $T'_{k-1} + t$. The event (i) has probability $p_c$. By the definition of the operator "Crossover with Repair" (Algorithm 2) event (ii) happens with probability at least $1/n^2$ and event (iii) happens again with probability at least $1/n^2$. Moreover, these bounds are independent of the event whether or not there already exists a weight-minimal approximating path of gap at most $g$ for the pair $(u, v)$ in the population at time $T'_{k-1} + t$. Thus, Equation (15) indeed holds for all $(u, v) \in I$ and $t \in \mathbb{N}$ with

$$p := \frac{(b(k-1)g(k))^2 p_c}{36n^4} = \frac{(5/4)^{2k} p_c (n \log n)^{4/5}}{81n^4}.$$

Finally, since the event that at time $T'_{k-1} + 1 + t$ one of the events $A^0_{(u,v)}, \ldots, A^t_{(u,v)}$ has occurred for all $(u, v) \in I$ implies the event "$T''_k \leq T'_{k-1} + 1 + t$", we have by Lemma 6 with $r := C_2(4/5)^{2k} n^3 (n \log n)^{0.2}$ and $C_2 := \frac{324}{p_c}$ that

$$pr = \frac{(5/4)^{2k} p_c (n \log n)^{4/5}}{81n^4} \cdot \frac{324}{p_c} \cdot (4/5)^{2k} n^3 (n \log n)^{0.2} = 4 \log n \geq 4 \ln n$$

and therefore, since $I \leq n^2$ that

$$\Pr\left[T''_k \geq T'_{k-1} + 1 + C_2(4/5)^{2k} n^3 (n \log n)^{1/5}\right] \leq |I| \mathrm{e}^{-4 \ln n} \leq \frac{1}{n^2}.$$

This concludes the proof of Proposition 13 and therefore also the proof of Theorem 1. □

## 6 Discussion

Apart from the results rigorously proven we conjecture that the bounds are actually tight and cannot be improved by a better analysis of the process.

In the following, we want to discuss how our runtime bounds can be extended to other but linear weight functions. For this, notice that the only place in the proofs of our runtime bounds where we refer to the actual properties of the weight-function is Lemma 3. In other words, if a fitness function $f$ on all paths of a graph $G$ satisfies Lemma 3, then our upper runtime bounds also hold for $f$. This motivates the following definition.

**Definition 16.** *Let $G$ be a finite, strongly connected directed graph and let $f$ be a non-negative fitness function on the set of all paths in $G$. Then $f$ is called* subpath optimal *if the following holds.*

*If $P_{u,v} = (u = u_0, \ldots, u_\ell = v)$ is an $f$-optimal $u$-$v$-path of length $\ell$ and $x = u_i$ and $y = u_j$ are vertices of $P_{u,v}$ with $0 \leq i \leq j \leq \ell$, then substituting the $f$-optimal subpath $P_{x,y}$ with another $f$-optimal path $P'_{x,y}$ yields an $f$-optimal path $P'_{u,v}$.*

Substituting a subpath $P_{x,y}$ of $P_{u,v}$ amounts to the concatenation of the three paths $P_{u,x}$ with $P'_{x,y}$ and $P_{y,v}$ at the vertices $x$ and $y$ (where one path can possibly be empty). Recall, that by design our crossover operators used in the Steady State $\text{GA}_{\text{APSP}}$ concatenate two or three paths to derive $P_{u,v}$. As in our analyses we ignored the ability of mutation to shrink paths by deleting edges, we regard the application of mutation as a concatenation of two paths.

We can now apply Lemma 3 to any weight-function with non-negative weights that is *subpath optimal*. This worked for the subpath optimal fitness function associated with APSP that maps paths to their lengths. Another subpath optimal example is mapping paths to the weight of their lightest edge (and maximize); this is known as the *all-pairs bottleneck paths* problem (see [13, 37]) and has applications for example in voting theory [33]. Lemma 3 is also applicable because if the minimum increases on a subpath then the overall minimum of the path is never decreased.

Thus, as a direct corollary to Theorem 1, the Steady State $\text{GA}_{\text{APSP}}$ with with repair on the all-pairs bottleneck paths problem has an optimization time of $O(n^{3.2}(\log n)^{0.2})$ iterations with high probability; and as a corollary to Theorem 2, we have that the Steady State $\text{GA}_{\text{APSP}}$ with feasible parent selection on the all-pairs bottleneck paths problem has an optimization time of $O(n^3 \log n)$ iterations with high probability.

As a final remark we would like to relate the success of our GA to dynamic programming. Our analysis is oriented at the Bellman-Ford Algorithm, and in particular the stages we consider are basically the same stages that occur there. From this perspective, one may say that the Steady State $\text{GA}_{\text{APSP}}$ mimics the optimization behavior of the Bellman-Ford Algorithm.

Still, the representation of individuals and the particular diversity mechanism used is natural since we would like to compute shortest paths for each pair of vertices. We see that the randomness introduced by the selection and crossover operators in the Steady State $\text{GA}_{\text{APSP}}$ raises subtle points in the analysis of our runtime bounds which are not present in that of the Bellman-Ford Algorithm. Two examples are the necessary overlap of the optimal subpaths in crossover and, in the case of Crossover with Repair, the interplay between mutation and crossover.

In a more general setting, one may ask whether our findings relate to other problems that have dynamic programming algorithms. The answer to this again relates to Lemma 3 and Definition 16. If the problem structure and representation within a GA is such that with sufficiently large probability the crossover operator can combine two optimal solutions like in Definition 16, we may hope that a GA using crossover and mutation outperforms a GA using mutation only. However, we are not aware of any concrete examples of such optimization problems, except for the APSP problems with subpath optimal fitness functions. We refer the reader to [6] for a discussion of how optimization problems which are accessible to dynamic programming can be solved by evolutionary algorithms with mutation only.

# 7  Conclusions

We have shown how the use of repair mechanisms or appropriate selection strategies can speed up crossover-based evolutionary algorithms for the all-pairs shortest path problem (and some other problems with a similar structure). However, it remains a challenge to understand the usefulness of crossover in evolutionary computation in a rigorous way for other combinatorial optimization problems.

The evolutionary algorithm examined for the APSP problem makes use of a strong diversity mechanism (each individual represents a path between a different pair of vertices) that allows to show the usefulness of crossover. Often evolutionary algorithms use much weaker diversity mechanism such as niching, deterministic crowding and fitness sharing and the goal is to compute just a single solution instead of a set of solutions. We state it as an open problem to show that crossover speeds up evolutionary algorithms for single-objective combinatorial optimization using one of the stated diversity mechanisms.

On the other hand, multi-objective problems use in a natural way a diversity mechanisms according to Pareto dominance. Often the population of an evolutionary algorithm for multi-objective optimization contains a population which represents the different tradeoffs with respect to the given objective function at a certain time step. It would be interesting to have rigorous results that show the usefulness of crossover in evolutionary multi-objective optimization.